%% file: root.tex
\documentclass[letterpaper, 10 pt, conference]{ieeeconf}  % Comment this line out if you need a4paper

\IEEEoverridecommandlockouts                              % This command is only needed if 
\usepackage{graphicx}
\usepackage{amsmath} % assumes amsmath package installed
\usepackage{amssymb}  % assumes amsmath package installed

\usepackage{array}
\usepackage{caption}
\usepackage{subcaption}
\usepackage{cite}
\usepackage{amsfonts}
\usepackage{algorithmic}
\usepackage{hyperref}
\usepackage{textcomp}
\usepackage{xcolor}
\usepackage{marvosym}
\usepackage{booktabs}
\usepackage{multirow}

\usepackage{titlesec}
\titlespacing{\subsection}{0pt}{6pt}{2pt}

\title{\LARGE \bf
REOcc: Camera-Radar Fusion with Radar Feature Enrichment \\for 3D Occupancy Prediction
}

\author{Chaehee Song$^{1}$, Sanmin Kim$^{2}$, Hyeonjun Jeong$^{3}$, Juyeb Shin$^{4}$, Joonhee Lim$^{4}$ and Dongsuk Kum$^{3}$% <-this % stops a space
\thanks{*This work was supported by Institute of Information and communications Technology Planning and Evaluation (IITP) and the National Research Foundation of Korea (NRF) grant funded by the Korea government (MSIT) (RS-2023-00236245, Development of Perception/Planning AI SW for Seamless Autonomous Driving in Adverse Weather/Unstructured Environment and 2022R1A2C2004944).}% <-this % stops a space
\thanks{$^{1}$C. Song is with the School of Electrical Engineering, Korea Advanced Institute of Science and Technology
(KAIST), Daejeon 34141, South Korea
        (email: {\tt\small chaehee.song@kaist.ac.kr})}%
\thanks{$^{2}$S. Kim is with the Department of Automobile and IT Convergence, Kookmin University, Seoul 02707, South Korea
        (email: {\tt\small sanmin.kim@kookmin.ac.kr})}%
\thanks{$^{3}$H. Jeong and D. Kum are with Cho Chun Shik Graduate School of Mobility, Korea Advanced Institute of Science and Technology
(KAIST), Daejeon 34141, South Korea
        (email: {\tt\small hyeonjun.jung, dskum @kaist.ac.kr})}%
\thanks{$^{4}$J. Shin and J. Lim are with the Robotics Program, Korea Advanced Institute of Science and Technology
(KAIST), Daejeon 34141, South Korea
        (email: {\tt\small juyebshin, kingear3 @kaist.ac.kr})}%
        }

\begin{document}

\maketitle
\thispagestyle{empty}
\pagestyle{empty}

\input{Section/0_abstract.tex}
\input{Section/1_introduction.tex}
\input{Section/2_related_work.tex}
\input{Section/3_methodology.tex}
\input{Section/4_experiments.tex}
\input{Section/5_conclusions.tex}
% \bibliographystyle{IEEEtran}
% \bibliography{Section/Ref}
\input{root.bbl}

\end{document}

%% file: Section/0_abstract.tex
% % 3d op task의 중요성
% % 기존 방법 문제 제기
% % 방법론 제안

% \begin{abstract}
% Vision-based 3D occupancy prediction has made significant advancements, addressing key challenges of traditional 3D object detection.
% Despite this, limitations of camera-only approaches have led to the development of sensor fusion methods.
% In particular, camera-radar fusion has shown great potential due to the complementary characteristics of the two sensors.
% However, the inherent sparsity and noise of the radar data pose challenges to achieving reliable performance.
% In this paper, we propose REOcc, a novel camera-radar fusion network designed to enrich radar feature representations for 3D occupancy prediction.
% Our approach introduces two main components, a Radar Densifier and a Radar Amplifier, which improve the spatial density and quality of the radar features by incorporating spatial and contextual information.
% Extensive experiments on Occ3D-nuScenes benchmark demonstrate the effectiveness of our model.
% REOcc achieves significant performance gains over the camera-only baseline model, particularly in dynamic object classes.
% These results underscore the ability of REOcc to effectively mitigate the limitations of raw radar data without reliance on additional sensors, unlocking its full potential for robust 3D occupancy prediction.
% % Experimental results demonstrate that our method improves 5.46 mIoU compared to the camera-only baseline model, proving the effectiveness of our approach.
% \end{abstract}

\begin{abstract}
Vision-based 3D occupancy prediction has made significant advancements, but its reliance on cameras alone struggles in challenging environments.
This limitation has driven the adoption of sensor fusion, among which camera-radar fusion stands out as a promising solution due to their complementary strengths.
However, the sparsity and noise of the radar data limits its effectiveness, leading to suboptimal fusion performance.
In this paper, we propose REOcc, a novel camera-radar fusion network designed to enrich radar feature representations for 3D occupancy prediction.
Our approach introduces two main components, a Radar Densifier and a Radar Amplifier, which refine radar features by integrating spatial and contextual information, effectively enhancing spatial density and quality.
Extensive experiments on the Occ3D-nuScenes benchmark demonstrate that REOcc achieves significant performance gains over the camera-only baseline model, particularly in dynamic object classes.
These results underscore REOcc's capability to mitigate the sparsity and noise of the radar data.
Consequently, radar complements camera data more effectively, unlocking the full potential of camera-radar fusion for robust and reliable 3D occupancy prediction.
\end{abstract}

%% file: Section/1_introduction.tex
\section{INTRODUCTION}
%% Extended Outline : Introduction %%
% 1. Background: 기존 연구 소개와 문제점
%     1-1. 기존 3D OD의 문제점
Accurate perception of the 3D surrounding scene is essential for autonomous driving systems, which must ensure reliability and safety.
% 3D object detection, a typical task in 3D perception, has drawn extensive attention and developed considerably over the past several decades. % playing a pivotal role in visual perception.
3D object detection, a typical task in 3D perception, has developed considerably over the past several decades.
% However, their traditional methods have inherent limitations that hinder the attainment of robust and reliable perception capabilities.
% The task aims to estimate the 3D position and dimensions of objects, primarily by generating bounding boxes for objects that belong to predefined categories.
% However, their methods focus on generating bounding boxes for objects belonging to predefined categories, meaning that objects outside these categories or with irregular geometries may exist but remain undetected.
However, their methods focus on generating cuboid bounding boxes for objects belonging to predefined categories, leaving out objects outside these categories and missing parts of irregularly shaped objects.
% implying that when faced with objects that lie beyond the predefined categories or possess irregular geometries, they exist yet remain undetected.
% It implies that the finite set of classes must be established in advance and objects that fall outside these categories remain undetectable.
% Furthermore, the rigid rectangular shape of the bounding box restricts its ability to accurately capture complex shapes, such as construction vehicles equipped with arms or ladder trucks, often neglecting fine details of these objects.
% It implies that when faced with objects that lie beyond the predefined categories or posses irregular geometries, they exist yet remain undetected. %, potentially leading to significant safety risks.

%     1-2. 3D OP의 등장
% 3D occupancy prediction has been proposed as a promising solution to overcome the limitations of 3D object detection.
3D occupancy prediction has emerged to address the growing demand for capturing dense and detailed 3D structures.
Beyond this, it serves as a promising solution to overcome the inherent limitations of 3D object detection.
% The innovative approach involves dividing the environment into voxels, determining the occupancy state of each tiny cube, and assigning semantic labels to the occupied voxels.For objects that are not in the predefined categories, they are labeled as General Objects(GOs).
% Notably, this approach does not depend on whether the objects conform to predefined classes or possess specific geometric details; the key factor is merely whether a tiny cube is occupied, irrespective of the object type.
% Notably, this approach does not depend on whether the objects conform to predefined classes or possess specific geometric details.
This approach determines the occupancy state of each voxel and assigns a semantic label, independent of object type. %independent of predefined object classes or specific geometric characteristics.
It not only enhances 3D modeling, but also seamlessly extends to various downstream tasks, including motion prediction and trajectory planning~\cite{tong2023scene}.
%     1-3. visioin-based 3D OP의 문제점: [환경 취약성, 3차원 정보 부족]으로 real-world에서 reliability, robustness 부족

The leading trend in 3D occupancy prediction is built on vision-based approaches because of cost-effectiveness and rich semantic information. % that leverage cameras as primary sensors.
% Although these approaches have shown promising results, relying on cameras remains a significant challenge.
% However, relying solely on cameras remains a significant challenge.
However, they rely on indirect spatial estimations due to the absence of explicit 3D information, and their performance is heavily influenced by environmental conditions, making them unreliable in challenging real-world scenarios.
% Moreover, their performance is heavily influenced by environmental conditions, such as low light, fog, or rain, making them less reliable in challenging real-world scenarios.
% In the context of autonomous driving, where safety is paramount, these limitations pose critical risks.

\begin{figure}[t]
    \centering
\includegraphics[width=0.97\linewidth]{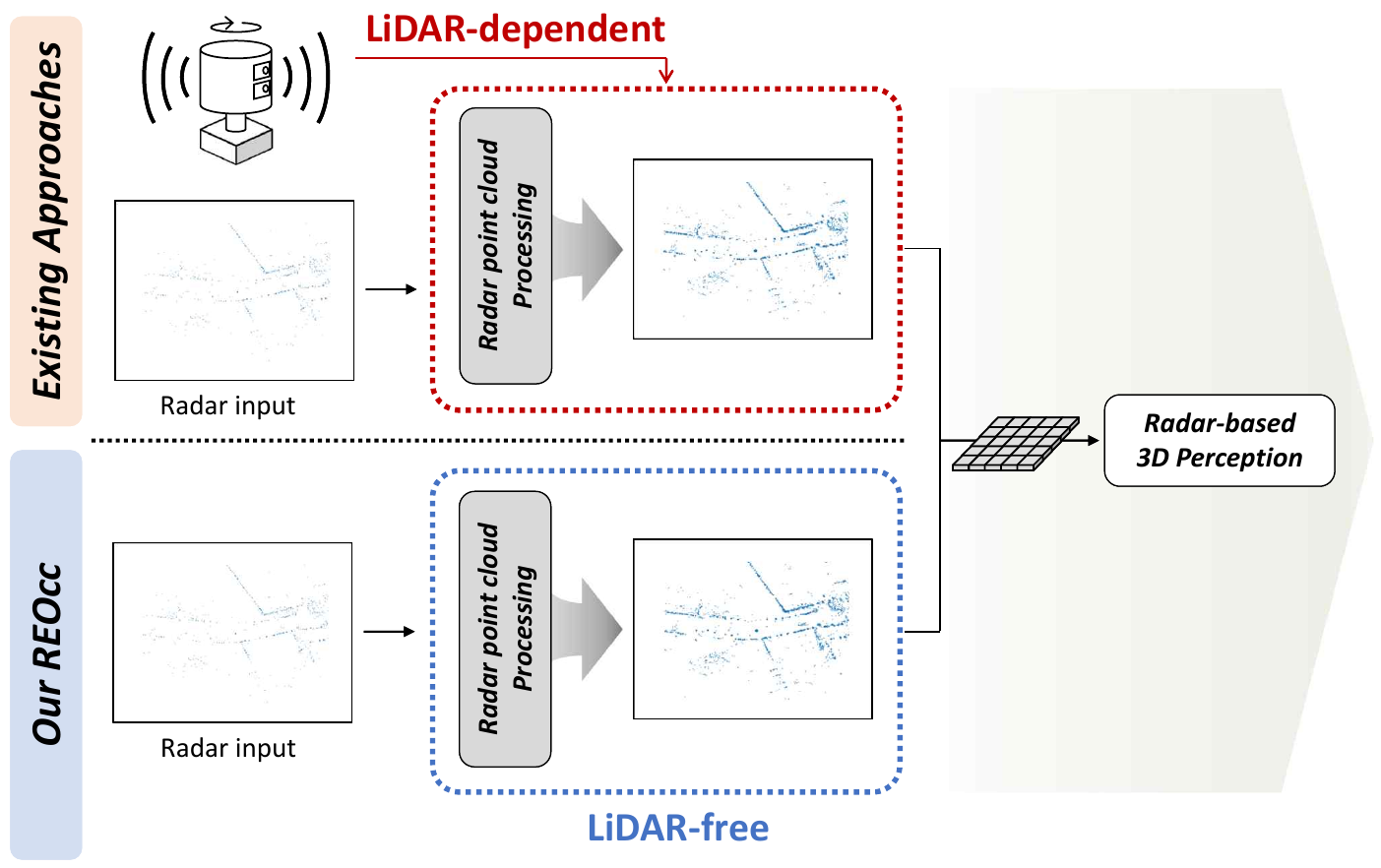}
    \vspace{-5pt}
    \caption{Illustration of our LiDAR-free REOcc. Existing approaches addressing radar's inherent sparse and noisy characteristics rely on LiDAR-based supervision. In contrast, our method achieves radar data processing using radar features alone, without supplementary sensors.}
    \label{fig:1_intro}
    \vspace{-15pt}
\end{figure}

\begin{figure*}[t]
    \centering
    \includegraphics[width=0.98\textwidth]{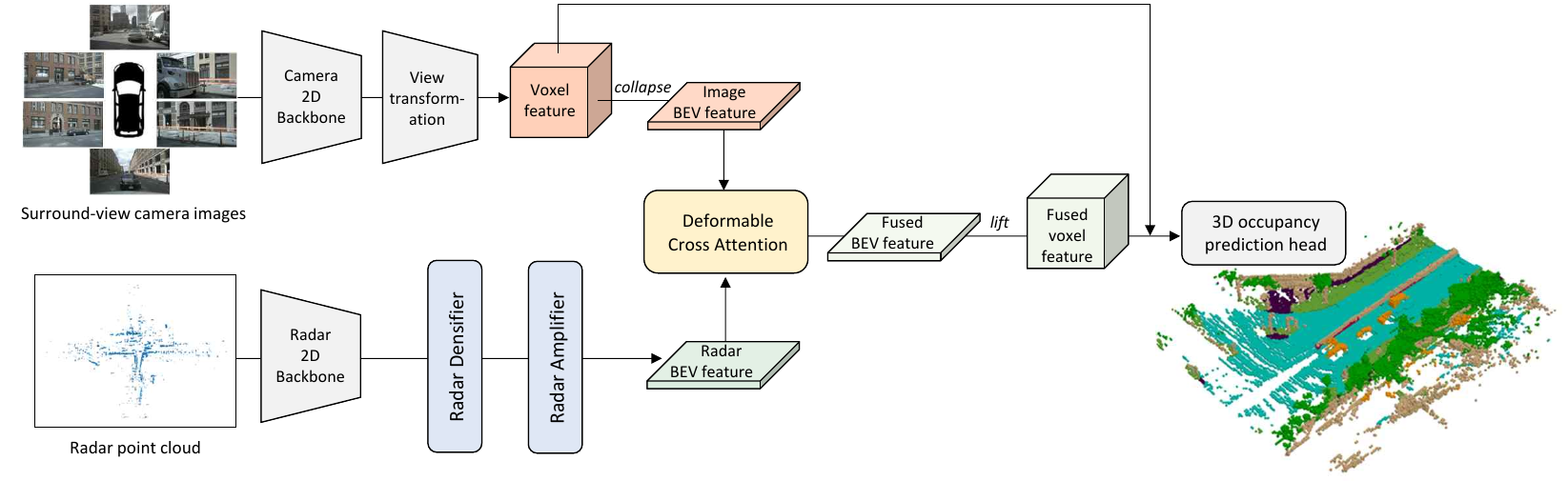}
    % \vspace{-10pt}
    \caption{Overall architecture of REOcc. The input data from multi-view camera images are processed through a 2D backbone and a view transformation to extract BEV features. In parallel, radar point clouds are fed into a separate 2D backbone to generate radar pillar features. Radar Densifier and Amplifier are then employed to improve both the quantity and quality of radar features, addressing the inherent sparse and noisy characteristics of radar data. Subsequently, the BEV features from the image and radar are fused and then lifted into the 3D volume with additional height information. Finally, an occupancy head is used to predict occupancy from the generated fused voxel features.}
    \label{fig:3_framework}
    \vspace{-15pt}
\end{figure*}
% 2. Motivation: 문제 제기
%     2-1. fusion 방법 등장 + 기존의 sensor fusion for 3D OP 방법의 문제점: 레이더 데이터의 단점을 해결하지 않고, 그대로 이용함.
The challenges lead to the development of multi-modal fusion strategies.
% mitigates the drawbacks of vision-based systems
% Especially radar sensors, which provide reliable depth information and exhibit robustness under varying environmental conditions, offer significant advantages over vision-only methods as complementary sensors, while also being more cost-effective than LiDAR.
In particular, radar sensors provide reliable depth information and maintain robustness under varying conditions.
They serve as complementary sensors to vision-based methods while being more cost-effective than LiDAR.
% Camera-radar fusion enables high performance at a relatively low cost, making it an attractive approach for autonomous driving systems where cost efficiency is a critical factor.
% Despite the promising potential of camera-radar fusion, previous methods have overlooked the issue of the radar's inherent limitations.
Despite the advantages of existing camera-radar fusion approaches, they have paid relatively less attention to the issue of the radar's inherent limitations.
% While radar offers advantages for dynamic 3D perception, its raw point clouds are typically sparse and cluttered, containing numerous false reflections and noise artifacts due to multi-path interference and specular reflections.
Radar's raw point clouds are typically sparse, making it challenging to extract dense and informative features. 
Additionally, they contain noisy artifacts, further complicating effective utilization.
Aforementioned sparsity and noise significantly degrade the reliability and accuracy of radar-based perception.
Simply fusing radar with other sensors without addressing these limitations can lead to suboptimal performance, as the fusion process may inherit the inherent weaknesses of radar rather than fully leveraging its strengths.
Therefore, effective radar processing strategies, such as denoising, feature enhancement, and spatial densification, are crucial to unveiling the full potential of radar for robust 3D perception in autonomous driving.

%     2-2. detection task에서 레이더 문제를 해결하고자 한 연구의 문제점: 학습과정에서 라이다의 도움이 필수적임 + training에서만 lidar를 사용하는 경우 문제점
% Several studies in other research fields have attempted to tackle these limitations.
To mitigate such issues, several works have explored the adoption of radar data processing techniques.
RadarDistill~\cite{bang2024radardistill}, which investigates 3D object detection through knowledge distillation from LiDAR to radar, introduces a Cross-Modality Alignment module to densify radar's low-level bird's-eye-view (BEV) features by leveraging multiple layers of dilation operations.
% They emphasize that the CMA plays a pivotal role in mitigating inefficiencies in knowledge transfer caused by the different densities of radar and LiDAR data, thereby enhancing the effectiveness of radar-based 3D object detection.
DenseRadar~\cite{han2024denserradar} introduces a 4D mmWave radar detector supervised by dense ground truth information generated from LiDAR point clouds.
% They first construct a dense 3D occupancy ground truth by stitching multiple frames of LiDAR point clouds.
% Using this ground truth for supervision, it extracts raw 4D mmWave radar features and refines them into radar features with higher density and accuracy, effectively densifying the detected radar point clouds.
Despite advancements in radar data processing, these methods remain dependent on LiDAR supervision during training. 
When LiDAR is used solely in the training phase, the model learns from data distributions inherently enriched with LiDAR information, creating a domain gap between training and inference.
%the trained model is exposed to data distributions inherently enriched with LiDAR information, creating a domain gap between training and inference. 
This discrepancy can cause significant performance degradation in real-world scenarios where LiDAR is unavailable~\cite{bang2024radardistill, shaw2023teacher}. 
Excessive reliance on LiDAR during training may also lead to biased feature representations that fail to generalize in LiDAR-free inference, compromising the robustness and reliability of radar-based perception.
Consequently, while the strategy using LiDAR only for training can reduce the storage and computational cost of the model, it increases the uncertainty during inference~\cite{sun2023monocular}.
Moreover, models trained with other types of data have no guarantee of generalization in different working conditions and domains~\cite{hu2022deep}.

% 3. Proposed method
%     3-1. Radar densifier, Radar amplifier 제안
%     3-2. 의의: 기존 카메라-레이더 퓨전 연구와 달리 레이더의 근본적인 문제점을 해결하고자 했으며,
%                 이를 위한 기존 연구와 달리, 다른 센서의 도움 없이 해결하고자 함.
In this paper, we propose REOcc, a novel camera-radar fusion network with radar feature enrichment for 3D occupancy prediction.
Unlike conventional approaches that rely on raw radar data or LiDAR-based supervision, our method enables the radar itself to tackle its inherent sparsity and noise, thereby maximizing the effectiveness of sensor fusion.
Specifically, we propose a Radar Densifier and Amplifier to enhance both the quantity and quality of radar features.
The Radar Densifier alleviates the sparsity by redistributing features to neighboring regions using distance-weighted sharing.
The Radar Amplifier further refines these features through probability-driven weighting, emphasizing informative channels while suppressing noise.
Notably, our method achieves radar data processing using radar features alone, without supplementary sensors like LiDAR, as illustrated in Fig. \ref{fig:1_intro}. 
This enrichment process generates more informative radar features, unveiling the potential of camera-radar fusion and ultimately enhancing the performance of fusion-based 3D perception systems for autonomous driving.
Our approach demonstrates notable performance gains, with a 15.02\% increase in mean Intersection over Union (mIoU) over the camera-only baseline. 
This improvement is particularly evident in dynamic object classes, achieving a 28.89\% increase, underscoring its effectiveness in detecting moving objects.
% Additionally, as a plug-and-play module, it easily integrates into existing 3D occupancy prediction methods.

%% file: Section/2_related_work.tex
\section{RELATED WORK}
% 1. Vision-based 3D OP
%     - 카메라 센서의 장점으로, 3D OP에서는 vision 중심으로 연구가 이루어져 왔음.
%     (rich semantic info, cost effectiveness, lateral accuracy)
%     - 관련 연구 소개 (surroundocc, scene as occupancy, occformer, occ3d,renderocc, fastocc, panoocc)
\subsection{Vision-based 3D Occupancy Prediction}
3D occupancy prediction task has drawn great attention due to its ability to provide intuitive and effective 3D scene understanding.
Early approaches to the task were primarily built on vision-based methods, largely due to the cost-effectiveness and rich semantic information offered by camera sensors.
% But they lack depth information and estimate 3D information from 2D features.
MonoScene~\cite{cao2022monoscene} pioneers the concept of occupancy prediction in outdoor environments using a camera.
% However, its reliance on a monocular camera proves insufficient for comprehensively capturing surround information.
However, its reliance on a monocular camera remains insufficient for capturing comprehensive surrounding information.
Subsequent works~\cite{tong2023scene,wei2023surroundocc,zhang2023occformer,tian2023occ3d,pan2024renderocc,hou2024fastocc,wang2024panoocc,huang2023tri,huang2024selfocc} adapt multi-view camera setups which demonstrate the benefits of aggregating multiple camera perspectives for enhanced 3D perception.
While most of these methods rely on BEV representations to describe 3D scenes, tri-perspective view (TPV) representations have recently been explored to further enrich spatial understanding~\cite{huang2023tri,zuo2023pointocc,huang2024selfocc}.
Despite these various advances in vision-based 3D occupancy prediction, challenges still exist due to the inherent lack of depth information and susceptibility to adverse environmental conditions.

\begin{figure}[t]
    \centering
    \includegraphics[width=0.9\linewidth]{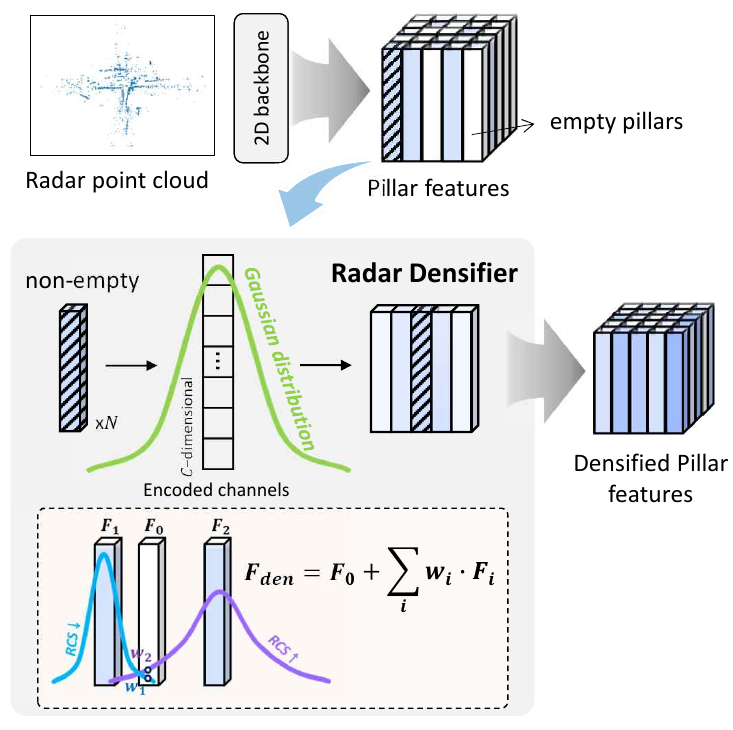}
    \vspace{-5pt}
    \caption{Detailed structure of the proposed Radar Densifier.}
    \label{fig:3_densifier}
    \vspace{-15pt}
\end{figure}

% 2. Sensor Fusion for 3D OP
%     - 하지만 카메라 센서의 한계(날씨/빛 조건 취약성)으로 더 신뢰도 있는 예측을 위해 다른 센서의 도움이 필요함.
%     - 카메라와 레이더는 상호보완적인 특징을 가지며, 퓨전을 통해 라이다에 준하는 성능을 낼 수 있다는 사실이 3D OD를 통해 증명됨.
%     - 관련 연구 소개

\subsection{Sensor Fusion for 3D Occupancy Prediction}
% Sensor fusion approaches for 3D occupancy prediction is proposed 
Several works on 3D occupancy prediction have integrated additional sensors with cameras to overcome the limitations of existing vision-based methods.
While LiDAR delivers highly accurate perception~\cite{bai2022transfusion,pang2020clocs,li2022deepfusion,xu2023fusionrcnn,wang2024occgen,pan2024co,vobecky2023pop,wang2023openoccupancy}, its high cost poses a significant barrier for large-scale deployment in autonomous vehicles.
In order to achieve LiDAR-free perception systems, the integration of alternative sensors is necessary.
Although lacking semantic information, radar provides reliable depth measurements and is resistant to low light and adverse weather conditions.
While radar falls behind LiDAR in performance, it exhibits complementary characteristics to cameras while also being a cost-effective sensor.
This makes camera-radar fusion an attractive approach, leveraging the complementary strengths of both sensors for improved 3D occupancy prediction.
This fusion strategy has been explored in other tasks, particularly in the 3D object detection task~\cite{kim2023crn,lei2023hvdetfusion,lin2024rcbevdet++,nabati2021centerfusion,kim2025crt,hwang2022cramnet,kim2023craft,kim2024rcm}.%, where it has demonstrated its effectiveness.
For example, Camera Radar Net (CRN)~\cite{kim2023crn} achieves performance comparable to LiDAR detectors and even outperforms them at longer distances, by addressing spatial misalignment between two inputs.

Early approaches~\cite{ming2024occfusion,wolters2024unleashing,lin2024teocc} have also applied camera-radar fusion to 3D occupancy prediction.
Hydra~\cite{wolters2024unleashing} employs a hybrid fusion approach that integrates radar and camera features at multiple stages, ensuring depth consistency through radar-guided alignment.
TEOcc~\cite{lin2024teocc} introduces a branch of temporal enhancement, drawing inspiration from the successful use of temporal information in 3D object detection~\cite{huang2022bevdet4d,li2023bevstereo,lin2022sparse4d,park2022time,wang2022sts}.
However, these methods do not consider the limitations inherent in raw radar point cloud data, such as its sparsity and clutter.
On the other hand, our approach involves radar data processing that addresses inherent limitations, enabling more effective camera-radar fusion and enhancing 3D occupancy prediction.

%% file: Section/3_methodology.tex
\section{METHODOLOGY}
\subsection{Overall Architecture}
The overall architecture of REOcc is shown in Fig. \ref{fig:3_framework}.
The framework is designed to enhance radar features and effectively integrate them with camera data for 3D occupancy prediction.
It consists of three main components: the Radar Densifier (\ref{sec:den}), Radar Amplifier (\ref{sec:amp}), and Camera-Radar Sensor Fusion (\ref{sec:fusion}).
The Radar Densifier addresses the sparsity of radar data by propagating features from non-empty pillars to nearby empty ones using a distance-based feature sharing.%, increasing spatial density.
The Radar Amplifier further refines radar features by selectively enhancing informative components and suppressing noise, improving the robustness of radar representations.
The Camera-Radar Sensor Fusion module then integrates enriched radar features with multi-view image features in the 2D BEV space, followed by a height refinement process to reconstruct 3D volumetric features.
By systematically enriching radar data and facilitating effective multi-modal fusion, REOcc achieves reliable 3D occupancy prediction.
The following sections provide detailed descriptions of each component.
% The overall framework is shown in Fig. \ref{fig:3_framework}.
% Multi-camera images are processed through a 2D backbone and a view transformation module to extract 3D volumetric features.
% Meanwhile, radar point clouds are fed into a separate 2D backbone to generate radar pillar features.
% These features are refined sequentially through Radar Densifier and Amplifier modules to mitigate sparsity and suppress noise.
% The refined radar features are then fused with image features in the 2D BEV domain.
% Subsequently, the fused 2D features are lifted into the 3D volume with additional height information. 
% To enhance height representation, the image-derived volumetric features are further concatenated with the lifted features in the height refinement branch.
% Finally, these refined features are employed for 3D occupancy prediction task.

\begin{figure}[t]
    \centering
    \includegraphics[width=0.95\linewidth]{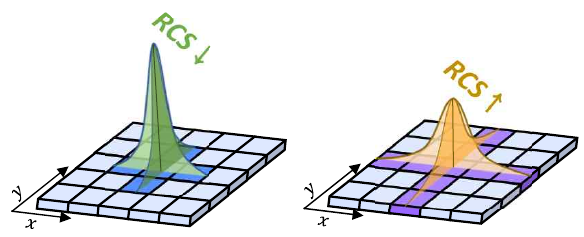}
    \vspace{-4pt}
    \caption{Illustration of RCS-based distribution. A Gaussian distribution reflecting the RCS is employed to account for the object's size, as the RCS provides a measure of the object's reflective surface area.}
    \label{fig:3_RCS}
    \vspace{-15pt}
\end{figure}

\subsection{Radar Densifier}
\label{sec:den}
The structure of the Radar Densifier is illustrated in Fig. \ref{fig:3_densifier}. 
Raw radar data processed through a point encoder with a 2D backbone and a neck~\cite{qi2017pointnet} often results in numerous empty pillars, occurring when no radar points exist within a particular section, as indicated in white.
This insufficiency of valid radar features poses challenges for effective fusion with camera data.
Radar point clouds also exhibit positional noise, which causes deviations from the actual location of the objects.

To address this issue, we propose the Radar Densifier, which enhances the spatial density of radar features through distance-based feature sharing.
We redistribute features from non-empty pillars to their neighbors using Gaussian-weighted function.
Here, the width of the Gaussian function, which determines the influence range of a radar point, is adaptively adjusted according to radar cross-section (RCS).
Since RCS reflects object size and radar reflectivity, higher values result in broader distributions, facilitating feature propagation over a wider area, as depicted in Fig. \ref{fig:3_RCS}.
By adaptively extending the spatial influence of radar features, the Radar Densifier not only alleviates sparsity but also addresses positional uncertainties, enhancing both the reliability and spatial consistency of radar feature representations.
The overall process is formulated as follows:
\begin{equation}
\label{eqn:densifier}
    F_{den}^{p}=F_{o}^{p}+\sum_{q \in \mathcal{N}(p)} w_{pq} F_{q}, 
\end{equation}
where $F_{den}^{p}$ and $F_{o}^{p}$ denote the densified and original features of pillar $p$, respectively.
$\mathcal{N}(p)$ represents the set of non-empty neighboring pillars that contribute to pillar $p$, and $F_{q}$ denotes the feature of a neighboring pillar $q \in \mathcal{N}(p)$.
$w_{pq}$ is a normalized Gaussian weight computed based on the spatial distance between the pillars $p$ and $q$. 

The densification process is applied for all non-empty pillars, with the Gaussian spreading performed along the $x$- and $y$-directions.
For computational efficiency, the redistribution is constrained within a predefined distance threshold using windowed filtering, ensuring that feature propagation remains both effective and efficient.

\begin{figure}[t]
    \centering
    \includegraphics[width=1.0\linewidth]{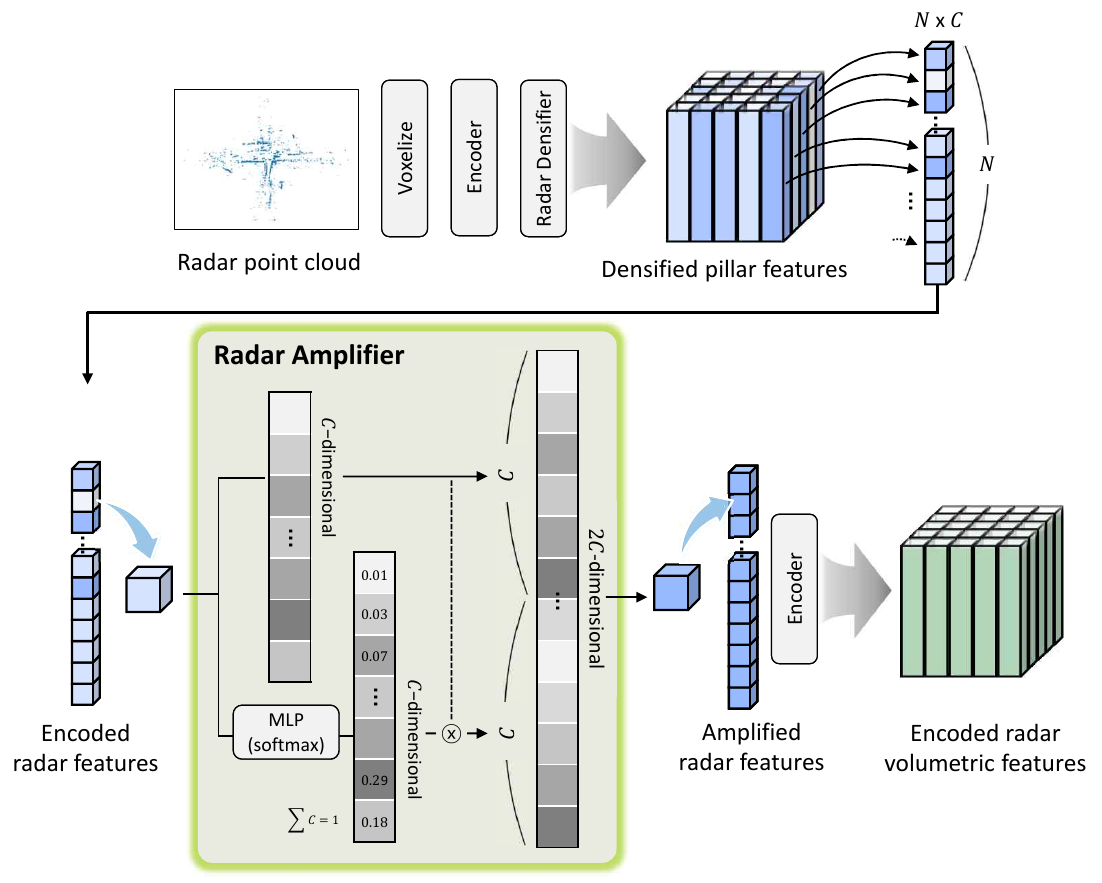}
    \caption{Detailed structure of the proposed Radar Amplifier.}
    \label{fig:3_amplifier}
    \vspace{-15pt}
\end{figure}

\subsection{Radar Amplifier}
\label{sec:amp}
Although the Radar Densifier alleviates the sparsity by redistributing distance-based features, radar measurements are still susceptible to noise, including false or weak reflections that lead to unreliable information.
To address this challenge, the Radar Amplifier is introduced, which selectively enhances meaningful radar features while suppressing noise.
The key idea is to estimate the relative importance of each feature channel and amplify those most relevant to the occupancy prediction, as illustrated in Fig. \ref{fig:3_amplifier}.

% The Radar Amplifier processes the densified pillar features $F_{den}^{rad}\in\mathbb{R}^{N\times C}$,
The Radar Amplifier enhances densified pillar features $F_{den}^{rad}\in\mathbb{R}^{N\times C}$ into amplified final radar features $F_{amp}^{rad}\in\mathbb{R}^{N\times C}$, where $N$ stands for the number of non-empty pillars, each encoded as a $C$-dimensional feature vector.
The module assigns a probability score to each feature channel, reflecting its contribution to the overall representation, and amplifies the features accordingly.
This process is formulated as follows:
\begin{equation}
\label{eqn:amplifier}
    F_{amp}=F_{den}^{rad}\oplus(\Phi_{prob}(F_{den}^{rad})\odot F_{den}^{rad}),
\end{equation}
\noindent
where $\Phi_{prob}$ is a multi-layer perceptron (MLP) with a softmax activation function, producing normalized probability scores across channels. %ensuring that the probability scores sum to one across channels.
The element-wise multiplication $\odot$ scales each feature according to its assigned probability, reinforcing informative channels while suppressing less relevant or noisy components.
The amplified features are then concatenated with the original densified features via $\oplus$ to maintain the full feature space while incorporating the importance-weighted representation.
This selective emphasis on high-confidence radar features reduces the impact of noise, improves feature robustness, and ultimately enhances the accuracy of 3D occupancy prediction.

\begin{table*}
\caption{3D occupancy prediction results on Occ3D-nuScenes validation set. Notion of modality: Camera (C), Radar (R). `const.veh.' and `drivable surf.' are the shorts for construction vehicles and drivable surfaces.}
\centering
\resizebox{\textwidth}{!}
    {
    \setlength{\tabcolsep}{1.5pt}
    \renewcommand{\arraystretch}{1.2}
    \begin{tabular}{l|c||c|ccccccccccccccccc}
    \hline
    Method & Input & mIoU$\uparrow$ & \rotatebox{90}{others} & \rotatebox{90}{barrier} & \rotatebox{90}{bicycle} & \rotatebox{90}{bus} & \rotatebox{90}{car} & \rotatebox{90}{const.veh.} & \rotatebox{90}{motorcycle} & \rotatebox{90}{pedestrian} & \rotatebox{90}{traffic cone} & \rotatebox{90}{trailer} & \rotatebox{90}{truck} & \rotatebox{90}{drivable surf.} & \rotatebox{90}{other flat} & \rotatebox{90}{sidewalk} & \rotatebox{90}{terrain} & \rotatebox{90}{manmade} & \rotatebox{90}{vegetation} \\
    \hline
    MonoScene~\cite{cao2022monoscene} & C & 6.06 & 1.75 & 7.23 & 4.26 & 4.93 & 9.38 & 5.67 & 3.98 & 3.01 & 5.90 & 4.45 & 7.17 & 14.91 & 6.32 & 7.92 & 7.43 & 1.01 & 7.65 \\
    BEVDet~\cite{huang2021bevdet} & C & 19.38 & 4.39 & 30.31 & 0.23 & 32.36 & 34.47 & 12.97 & 10.34 & 10.36 & 6.26 & 8.93 & 23.65 & 52.27 & 24.61 & 26.06 & 22.31 & 15.04 & 15.10 \\
    BEVFormer~\cite{li2024bevformer} & C & 26.88 & 5.85 & 37.83 & 17.87 & 40.44 & 42.43 & 7.36 & 23.88 & 21.81 & 20.98 & 22.38 & 30.70 & 55.35 & 28.36 & 36.00 & 28.06 & 20.04 & 17.69 \\
    BEVStereo~\cite{li2023bevstereo} & C & 24.51 & 5.73 & 38.41 & 7.88 & 38.70 & 41.20 & 17.56 & 17.33 & 14.69 & 10.31 & 16.84 & 29.62 & 54.08 & 28.92 & 32.68 & 26.54 & 18.74 & 17.49 \\
    TPVFormer~\cite{huang2023tri} & C & 27.83 & 7.22 & 38.90 & 13.67 & 40.78 & 45.90 & 17.23 & 19.99 & 18.85 & 14.30 & 26.69 & 34.17 & 55.65 & 35.47 & 37.55 & 30.70 & 19.40 & 16.78 \\
    OccFormer~\cite{zhang2023occformer} & C & 21.93 & 5.94 & 30.29 & 12.32 & 34.40 & 39.17 & 14.44 & 16.45 & 17.22 & 9.27 & 13.90 & 26.36 & 50.99 & 30.96 & 34.66 & 22.73 & 6.76 & 6.97 \\
    CTF-Occ~\cite{tian2023occ3d} & C & 28.53 & 8.09 & 39.33 & 20.56 & 38.29 & 42.24 & 16.93 & 24.52 & 22.72 & 21.05 & 22.98 & 31.11 & 53.33 & 33.84 & 37.98 & 33.23 & 20.79 & 18.00 \\
    SurroundOcc~\cite{wei2023surroundocc} & C & 37.18 & 8.97 & 46.33 & 17.08 & 46.54 & 52.01 & 20.05 & 21.47 & 23.52 & 18.67 & 31.51 & 37.56 & 81.91 & 41.64 & 50.76 & 53.93 & 42.91 & 37.16 \\
    FastOcc~\cite{hou2024fastocc} & C & 39.21 & 12.06 & 43.53 & 28.04 & 44.80 & 52.16 & 22.96 & 29.14 & 29.68 & 26.98 & 30.81 & 38.44 & 82.04 & 41.93 & 51.92 & 53.71 & 41.04 & 35.49 \\
    \hline
    OccFusion~\cite{ming2024occfusion} & C+R & 38.26 & 10.11 & 43.48 & 26.03 & 48.25 & 53.29 & 24.98 & 33.35 & 31.75 & 27.31 & 29.88 & 40.15 & 77.59 & 35.47 & 43.64 & 48.01 & 42.73 & 33.81 \\
    TEOcc~\cite{lin2024teocc} & C+R & 42.90 & 10.82 & 50.33 & 24.28 & 48.99 & 57.32 & 29.38 & 24.41 & 30.14 & 28.46 & 36.46 & 43.01 & 83.96 & 43.09 & 56.00 & 59.34 & 54.18 & 49.16 \\ 
    REOcc (Ours) & C+R & \textbf{45.33} & 13.32 & 51.53 & 31.82 & 52.46 & 58.71 & 29.55 & 32.78 & 32.99 & 32.08 & 39.92 & 44.88 & 84.54 & 46.44 & 57.69 & 60.32 & 54.04 & 47.50 \\
        \hline
    \end{tabular}
    }
    \label{tab:4_OccMain}
\vspace{-3pt}
\end{table*}

\begin{table*}
\caption{Comparison with the state-of-the-art method for 3D occupancy prediction. TEOcc employs its own single-modal TEOcc model as the camera-only baseline, utilizing ResNet-50 as the backbone network. We use BEVDet4D-Occ as a baseline for camera-only baseline. To ensure a fair comparison, REOcc is also trained with ResNet-50 as its backbone.}
\centering
\resizebox{\textwidth}{!}
    {
    \setlength{\tabcolsep}{1pt}
    \renewcommand{\arraystretch}{1.2}
    \begin{tabular}{l|c|c||c|ccccccccccccccccc}
    \hline
    Method & Backbone & Input & mIoU$\uparrow$ & \rotatebox{90}{others} & \rotatebox{90}{barrier} & \rotatebox{90}{bicycle} & \rotatebox{90}{bus} & \rotatebox{90}{car} & \rotatebox{90}{const.veh.} & \rotatebox{90}{motorcycle} & \rotatebox{90}{pedestrian} & \rotatebox{90}{traffic cone} & \rotatebox{90}{trailer} & \rotatebox{90}{truck} & \rotatebox{90}{drivable surf.} & \rotatebox{90}{other flat} & \rotatebox{90}{sidewalk} & \rotatebox{90}{terrain} & \rotatebox{90}{manmade} & \rotatebox{90}{vegetation} \\
    \hline
    TEOcc-baseline & ResNet50 & C & 39.36 & 9.59 & 47.60 & 13.82 & 43.91 & 52.87 & 27.92 & 17.58 & 23.89 & 21.69 & 33.91 & 39.60 & 83.38 & 41.84 & 54.94 & 57.92 & 50.83 & 47.23 \\ 
    TEOcc & ResNet50 & C+R & 42.90 & 10.82 & 50.33 & 24.28 & 48.99 & 57.32 & 29.38 & 24.41 & 30.14 & 28.46 & 36.46 & 43.01 & 83.96 & 43.09 & 56.00 & 59.34 & 54.18 & 49.16 \\ 
    \hline
    REOcc-baseline & ResNet50 & C & 36.34 & 8.39 & 44.47 & 14.84 & 40.29 & 49.35 & 21.74 & 20.23 & 21.09 & 21.68 & 28.30 & 35.30 & 80.85 & 39.65 & 52.20 & 54.92 & 45.03 & 39.50 \\ 
    REOcc & ResNet50 & C+R & \textbf{41.80} & 10.22 & 48.52 & 17.57 & 47.84 & 57.81 & 29.78 & 24.58 & 28.15 & 26.70 & 35.33 & 41.71 & 83.64 & 42.38 & 56.03 & 58.88 & 54.73 & 46.70 \\
    \hline    
    REOcc-baseline & SwinT & C & 42.02 & 12.15 & 49.63 & 25.10 & 52.02 & 54.46 & 27.87 & 27.99 & 28.94 & 27.23 & 36.43 & 42.22 & 82.31 & 43.29 & 54.62 & 57.90 & 48.61 & 43.55 \\ 
    REOcc & SwinT & C+R & \textbf{45.33} & 13.32 & 51.53 & 31.82 & 52.46 & 58.71 & 29.55 & 32.78 & 32.99 & 32.08 & 39.92 & 44.88 & 84.54 & 46.44 & 57.69 & 60.32 & 54.04 & 47.50 \\
        \hline
    \end{tabular}
    }
    \label{tab:4_OccSOTA}
\vspace{-10pt}
\end{table*}

% \begin{table*}[tb]
% \caption{3D occupancy prediction results for dynamic classes on Occ3D-nuScenes validation set. In addition to the mIoU based on 17 different classes, we also highlight the mIoU for the 8 dynamic classes: bicycle, bus, car, construction vehicle, motorcycle, pedestrian, trailer, and truck.}
%     \centering
%     \resizebox{0.82\textwidth}{!}
%     {
%     \setlength{\tabcolsep}{4pt}
%     \renewcommand{\arraystretch}{1.2}
%     \begin{tabular}{l|c|c||c|cccccccc}
%     % {>{\raggedright\arraybackslash}m{3cm}||>{\centering\arraybackslash}m{2cm}||>{\centering\arraybackslash}m{2cm}|*{8}{c}}  
%     \hline
%         Method & Backbone & Input & mIoU$\uparrow$ & \rotatebox{90}{bicycle} & \rotatebox{90}{bus} & \rotatebox{90}{car} & \rotatebox{90}{const.veh.} & \rotatebox{90}{motorcycle} & \rotatebox{90}{pedestrian} & \rotatebox{90}{trailer} & \rotatebox{90}{truck} \\
%         \hline
%         REOcc-baseline & ResNet50 & C & 28.89 & 14.84 & 40.29 & 49.35 & 21.74 & 20.23 & 21.09 & 28.30 & 35.30 \\ 
%         REOcc-baseline & SwinT & C & 36.88 & 25.10 & 52.02 & 54.46 & 27.87 & 27.99 & 28.94 & 36.43 & 42.22 \\ 
%         \hline
%         REOcc & ResNet50 & C+R & \textbf{35.35} & 17.57 & 47.84 & 57.81 & 29.78 & 24.58 & 28.15 & 35.33 & 41.71 \\
%         REOcc & SwinT & C+R & \textbf{40.39} & 31.82 & 52.46 & 58.71 & 29.55 & 32.78 & 32.99 & 39.92 & 44.88 \\
%         \hline
%     \end{tabular}
%     }
%     \label{tab:4_OccDynamic}
% \end{table*}

\subsection{Camera-Radar Sensor Fusion}
\label{sec:fusion}
While radar point clouds are refined sequentially through the proposed Radar Densifier and Amplifier, image features are extracted from multi-view cameras using an image backbone (e.g., ResNet~\cite{he2016deep}, SwinTransformer~\cite{liu2021swin}) with an additional FPN~\cite{lin2017feature} to capture multi-scale feature representations.
These features are then passed through a view transformation module to generate 3D volumetric features.
At this stage, we have two groups of features: volumetric image features $F_{3D}^{img}\in\mathbb{R}^{C_I\times Z_I\times H\times W}$ and 2D radar features $F_{2D}^{rad}\in\mathbb{R}^{C_R\times H\times W}$.
Since $F_{3D}^{img}$ encodes height information in a voxelized structure, it is collapsed into a 2D BEV representation by multiplying the channel and height dimensions, yielding $F_{2D}^{img}\in\mathbb{R}^{(C_I*Z_I)\times H\times W}$.
Following the multi-modal deformable cross attention from ~\cite{kim2023crn}, 2D features from both modalities are fused.
Specifically, the concatenated 2D features act as a query, while both image and radar features serve as keys and values, allowing the network to learn cross-modal dependencies. %enabling effective cross-modal feature interaction.
However, the resulting fused representation remains in the 2D domain, which limits its direct applicability for 3D occupancy prediction.

To bridge this gap, we adopt a height re-projection operation inspired by \cite{yu2023flashocc}.
The fused 2D features, initially represented as ${C_{fused}\times H\times W}$, are reshaped into ${C^*_{fused}\times Z\times H\times W}$, where $Z$ denotes the height dimension in 3D space.
Here, ${C_{fused}}$ and $C^*_{fused}$ denote the number of channels in the 2D and 3D space, respectively, satisfying the relation $C_{fused}=C^*_{fused}\times Z$.
While this transformation reintroduces height information, it lacks explicit supervision, potentially leading to inaccuracies.
To further refine the reconstructed height information, the original volumetric image features $F_{3D}^{img}$ are concatenated with the reshaped fused features.
This enhances the quality of the 3D-transformed features for 3D occupancy prediction.

%% file: Section/4_experiments.tex
\section{EXPERIMENTS}
\subsection{Dataset and Metrics}
We conduct experiments on Occ3D-nuScenes~\cite{tian2023occ3d}, which is built on the nuScenes dataset~\cite{caesar2020nuscenes}.
It provides 3D ground-truth labels using voxels, including 16 semantic classes and a free class.
We follow the evaluation metric used in the Occ3D-nuScenes benchmark, the mIoU score, which is computed by averaging the IoU values across all classes.

\subsection{Implementation Details}
For the camera stream, we adopt BEVDet4D~\cite{huang2022bevdet4d} as a baseline.
Its efficient architecture makes it well-suited for sensor fusion experiments, as it avoids the excessive computational overhead often associated with state-of-the-art (SOTA) models.
It comprises the feature extraction, view transformation, and feature encoding process.
We accumulate features from the eight previous image frames and concatenate them with the features of the current frame.
For radar, we employ PointPillar~\cite{shi2022pillarnet} to extract features in each pillar.
% For radar, we employ PointPillar~\cite{shi2022pillarnet} to divide radar points into pillars and use PointNet~\cite{qi2017pointnet} to extract features in each pillar.
Our models are trained for 24 epochs using the AdamW~\cite{loshchilov2017decoupled} optimizer with an initial learning rate of 4e-4 and a weight decay of 1e-2. 
We use a batch size of 4 on 8 NVIDIA RTX A6000 GPUs.

\begin{table}[tb]
\caption{Performance improvement of the camera-radar fusion over camera-only baseline for 3D occupancy prediction. mIoU and $\text{mIoU}_d$ represent the mean IoU for all classes and the subset of 8 dynamic classes, respectively. $\Delta_{\text{mIoU}}$ and $\Delta_{\text{mIoU}_d}$ denote the corresponding performance gains in mIoU achieved by radar integration.}
    \centering
    \resizebox{\columnwidth}{!}
    {
    \setlength{\tabcolsep}{3pt}
    \renewcommand{\arraystretch}{1.2}
    \begin{tabular}{l||cc|cc}
    \hline
        \multirow{2}{*}{Method} & \multicolumn{2}{c|}{All classes} & \multicolumn{2}{c}{Dynamic classes} \\
        & mIoU$\uparrow$ & $\Delta_{\text{mIoU}}$$\uparrow$ & $\text{mIoU}_d$$\uparrow$ & $\Delta_{\text{mIoU}_d}$$\uparrow$ \\
        \hline
        TEOcc & 42.90 & 3.54 (8.99\%) & 36.75 & 5.06 (15.97\%) \\ 
        REOcc & 41.80 & \textbf{5.46 (15.02\%)} & 35.35 & \textbf{6.46 (22.36\%)} \\ 
        \hline
    \end{tabular}
    }
    \label{tab:4_OccComparison}
    \vspace{-20pt}
\end{table}

\begin{figure*}[t]
    \centering
    \includegraphics[width=0.95\textwidth]{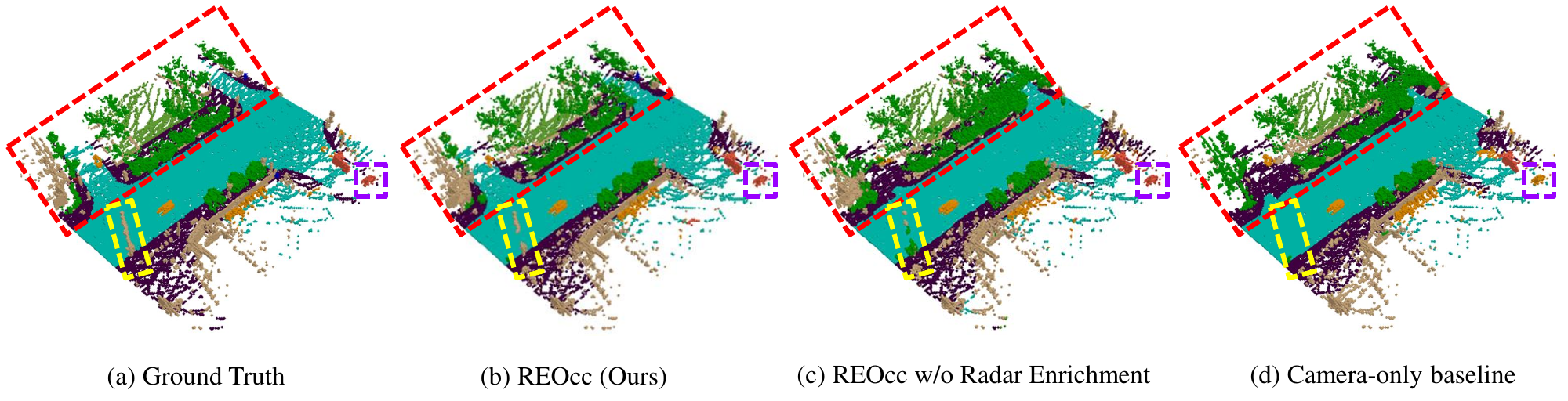}
    \vspace{-5pt}
    \caption{Visualization of 3D occupancy prediction results. Our REOcc generates more precise occupancy predictions compared to REOcc without radar enrichment and the camera-only baseline model.}
    \label{fig:4_results}
    \vspace{-15pt}
\end{figure*}

\subsection{Main Results}
%We compare REOcc with the camera-only method for 3D occupancy prediction on the Occ3D-nuScenes validation set, as shown in \tableautorefname~\ref{tab:4_OccMain}.
We compare REOcc with previous 3D occupancy prediction methods~\cite{cao2022monoscene,huang2021bevdet,li2024bevformer,li2023bevstereo,huang2023tri,zhang2023occformer,tian2023occ3d,wei2023surroundocc,hou2024fastocc
,ming2024occfusion,lin2024teocc} on the Occ3D-nuScenes validation set, as listed in \tableautorefname~\ref{tab:4_OccMain}.
Our REOcc achieves 45.33 mIoU, outperforming all previous methods.
This performance gain stems from effectively addressing the inherent sparsity and noise of radar data, allowing our model to fully leverage radar information for dense prediction in the occupancy task.
Consequently, our model benefits from radar more effectively than other fusion methods like OccFusion~\cite{ming2024occfusion}, achieving greater performance.
\tableautorefname~\ref{tab:4_OccSOTA} illustrates the comparison of the mIoU scores under the same backbone network with TEOcc~\cite{lin2024teocc}, a state-of-the-art camera-radar fusion model for 3D occupancy prediction.
The slightly lower performance of our REOcc may be influenced by differences in the choice of the camera-only baseline.
To balance computational efficiency and sensor fusion compatibility, we adopted BEVDet4D~\cite{huang2022bevdet4d} as the baseline model.
% To ensure computational efficiency and facilitate sensor fusion, we adopted BEVDet4D as the baseline.
However, employing a more advanced camera-only perception model specifically optimized for perception performance could further enhance our radar-fusion approach. 
Although our REOcc falls slightly behind TEOcc in the final evaluation, it delivers significantly greater gains over the camera-only baseline.
This will be further detailed in \tableautorefname~\ref{tab:4_OccComparison}.
% Our REOcc achieves mIoU of 41.80 with a ResNet-50 backbone, outperforming the camera-only baseline by 15.02\%.
% With a SwinTransformer backbone, it attains mIoU of 45.33, yielding a 7.88\% improvement.
% These results demonstrate the effectiveness of our approach in leveraging radar data for enhanced 3D occupancy prediction.
% The performance gains are primarily attributed to the integration of radar data, which complements camera images with additional spatial and motion cues.
% In addition, our method effectively refines and utilizes radar features through the Radar Densifier and Radar Amplifier, ensuring more robust and informative representations.
%As shown in \tableautorefname~\ref{tab:4_OccDynamic}, 

We conduct an analysis of our model on dynamic object classes to further evaluate the impact of radar fusion.
Camera-only perception often struggles with detecting and tracking dynamic objects.
In contrast, radar sensors provide depth and velocity measurements while enabling object detection at relatively long distances (more than 200 meters), which is crucial for the robust detection of moving objects~\cite{cho2014multi}.
Given these complementary characteristics, assessing performance specifically on dynamic objects offers a more meaningful evaluation of radar integration in real-world scenarios.
To this end, we define a subset of dynamic object classes, including bicycle, bus, car, construction vehicle, motorcycle, pedestrian, trailer, and truck, selected from the original 17 classes.
Within these dynamic object classes, our model achieves mIoU of 35.35 with the ResNet-50 backbone and 40.39 with the SwinTransformer backbone, significantly outperforming the camera-only baseline by 6.46 and 3.51 mIoU, respectively.
These results highlight the effectiveness of radar fusion in improving dynamic object perception.
The significant performance gains demonstrate that our model successfully mitigates camera-only perception limitations by effectively leveraging radar data, leading to more robust 3D occupancy predictions.

\tableautorefname~\ref{tab:4_OccComparison} provides a comparative analysis of the improvement ($\Delta$) achieved over each model’s baseline.
Across all classes, TEOcc exhibits 3.54 mIoU (8.99\%) increase, while our REOcc surpasses this with a higher gain of 5.46 mIoU (15.02\%).
The advantage becomes even more evident in dynamic object classes, where TEOcc improves by 5.06 mIoU (15.97\%) increase, whereas REOcc outperforms it with a substantial gain of 6.46 mIoU (22.36\%).
Since this study focuses on the sensor fusion strategy for 3D occupancy prediction, these findings strongly validate the effectiveness of our radar-camera fusion method. 
The significant improvement observed in REOcc underscores its ability to better leverage radar data, particularly in enhancing the perception of dynamic objects. 
These results highlight the strength of our approach in effectively enriching radar data and seamlessly integrating it with camera inputs, ultimately leading to a more precise and reliable multi-modal fusion framework.

\begin{table}
\caption{Ablation studies for evaluating the contribution from individual component of radar enrichment process on Occ3D-nuScenes validation set.}
\vspace{-3pt}
    \label{tab:4_Abl}
\centering
\resizebox{\columnwidth}{!}
    {
    \setlength{\tabcolsep}{4pt}
    \renewcommand{\arraystretch}{1.2}
    \begin{tabular}{cc|cc}
        \hline
        Radar Densifier & Radar Amplifier & mIoU$\uparrow$ & $\text{mIoU}_d$$\uparrow$ \\
        \hline
         &  & 39.43  & 32.20 \\
        \checkmark &  & 40.68 & 34.19 \\
         & \checkmark & 40.70 & 33.92 \\
        \checkmark & \checkmark & \textbf{41.80} & \textbf{35.35} \\
        \hline
    \end{tabular}
    }
\vspace{-15pt}
\end{table}

\begin{figure*}[t]
    \centering
    \includegraphics[width=0.95\textwidth]{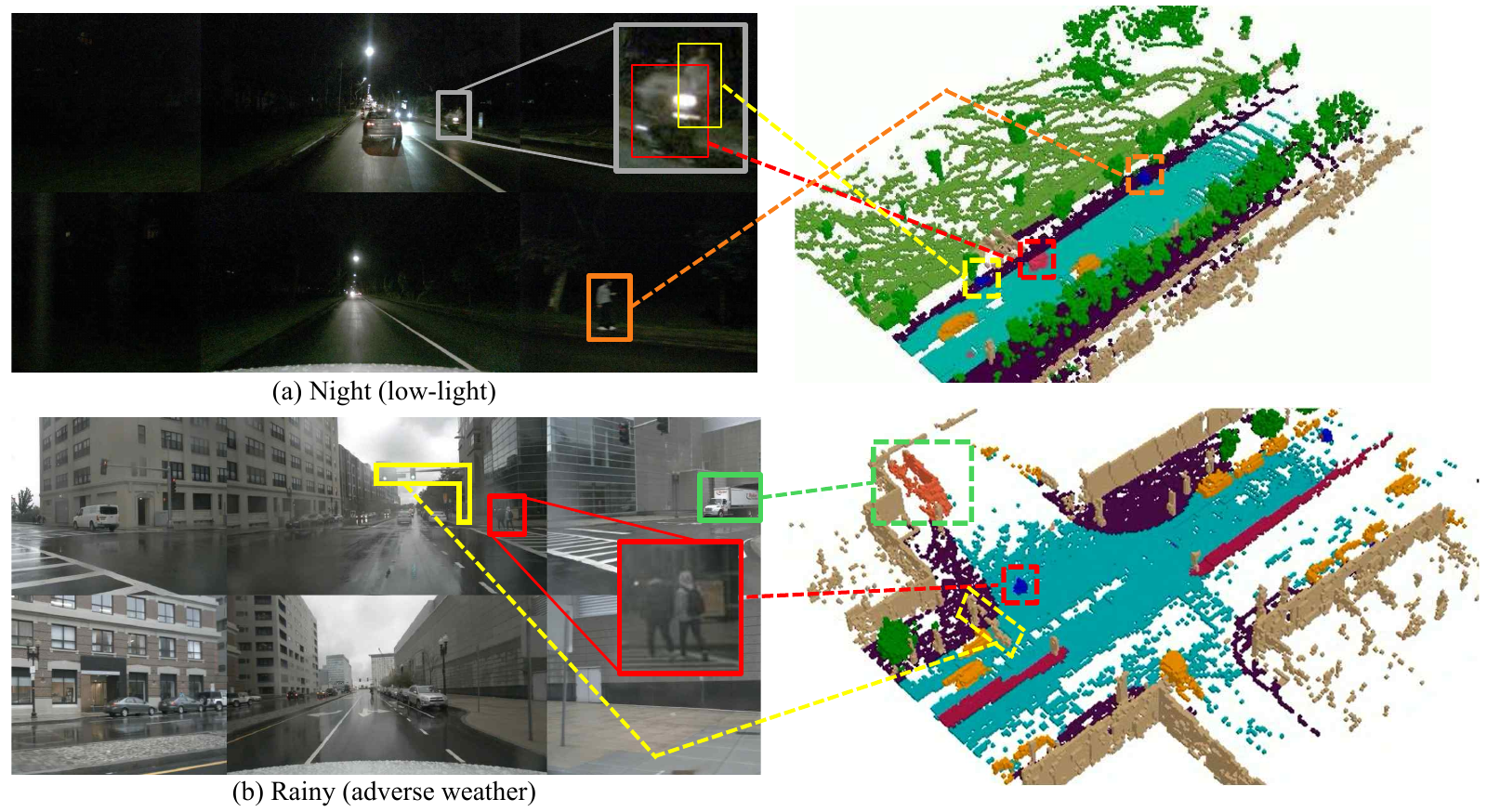}
    \caption{Visualization of 3D occupancy prediction results on challenging scenarios. We show 6 camera views surrounding the vehicle and our prediction results for each of the (a) night and (b) rainy scene. It demonstrates the capability of our model to detect and classify objects that are difficult to perceive from camera images alone.}
    \label{fig:4_results2}
    \vspace{-17pt}
\end{figure*}

\subsection{Ablation Study}
We conduct ablation studies on the Occ3D-nuScenes validation set to evaluate the contribution of each component in the radar enrichment process.
All experiments employ ResNet-50 as the backbone network and are trained for 24 epochs.
\tableautorefname~\ref{tab:4_Abl} presents the performance gains achieved by individual radar processing modules.
Integrating the Radar Densifier into the camera-radar fusion network leads to the mIoU improvement of 1.25 for all classes and 1.99 for 8 dynamic classes, demonstrating its effectiveness in mitigating radar sparsity.
Similarly, incorporating only the Radar Amplifier results in gains of 1.27 and 1.72 mIoU for all classes and dynamic classes, respectively, highlighting its critical role in the refinement of radar features.
% In contrast, when neither the Radar Densifier nor the Radar Amplifier is applied—excluding the radar enrichment process—the model results in an 39.43 mIoU for all classes and 32.20 mIoU for dynamic classes.
Without both radar enrichment modules, the model achieves 39.43 mIoU for all classes and 32.20 mIoU for dynamic classes, indicating the effectiveness of our fusion strategies.
Nevertheless, removing radar enrichment from REOcc results in a 5.67\% decrease in mIoU for the fusion network.
In particular, the perception of dynamic objects declines by 8.91\%, highlighting the crucial role of radar enrichment in detecting moving objects.
% These results highlight that refining and effectively utilizing radar information is crucial for improving 3D occupancy prediction performance.

\subsection{Qualitative Results}% under Challenging Conditions}
Fig. \ref{fig:4_results} shows the qualitative visualization results of REOcc compared with ours without radar enrichment and the camera-only baseline model.
Several static object classes, including man-made structures (indicated in yellow) and surfaces such as drivable areas and sidewalks (indicated in red), are accurately detected by REOcc.
Additionally, the fact that the camera-only model fails to predict dynamic object classes (indicated in purple) underscores the contribution of radar fusion to dynamic object perception.

In Fig. \ref{fig:4_results2}, we present qualitative results under challenging conditions such as night and rain.
% Camera image data often lose critical information in these scenarios due to poor illumination and visual obstructions.
As seen in the provided examples, camera image data in these scenarios can be affected by factors such as low illumination and occlusions, potentially leading to missing critical details.
However, REOcc, which integrates well-calibrated radar data, demonstrates strong robustness in such adverse environments.
In the night scene, our model successfully detects a motorcycle and pedestrians that are barely visible in the images, even distinguishing two overlapping objects.
Similarly, in rainy conditions, REOcc identifies pedestrians, a traffic light, and a truck, despite their blurred appearance in the images. 
Furthermore, our model accurately captures the structure of the crossbar of the traffic sign that extends over the road (indicated by the yellow line), demonstrating its ability to estimate the heights of objects.

%% file: Section/5_conclusions.tex
\section{CONCLUSIONS}
In this paper, we propose REOcc, a novel camera-radar fusion network for 3D occupancy prediction, designed to maximize the utility of radar data through an enrichment process.
To achieve this, REOcc consists of two main components: Radar Densifier and Radar Amplifier.
The Radar Densifier mitigates the sparsity of radar data by applying distance-based feature sharing.
The Radar Amplifier further refines radar features by selectively enhancing informative components and suppressing noise.
These modules improve the spatial density and quality of the radar features, facilitating a more effective sensor fusion.
Extensive experiments on the Occ3D-nuScenes validation set demonstrate the efficacy of REOcc for 3D occupancy prediction.
These findings underscore that our REOcc effectively addresses the inherent sparsity and noise of raw radar data without reliance on supplementary sensors. 
In addition, our approach exhibits substantial performance gains over its camera-only baseline, particularly in dynamic object classes, underscoring the importance of properly refining and leveraging radar data for effective radar fusion.
Independently addressing the fundamental limitations of raw radar data, REOcc establishes a new paradigm for radar processing in radar-based perception frameworks.
This work serves as a foundation for future research in accurate and reliable 3D scene perception, paving the way for more sophisticated sensor fusion strategies.